\documentclass[conference]{IEEEtran}
\IEEEoverridecommandlockouts
\usepackage[table,svgnames]{xcolor}
\usepackage{multirow}
\usepackage{pdfpages}
\usepackage{graphicx}
\usepackage{float}
\usepackage{amsmath}
\usepackage{threeparttable}
\usepackage[utf8]{inputenc}
\usepackage[english]{babel}
\usepackage{fancyhdr}
\usepackage{lastpage}
\usepackage{caption}
\usepackage{tabulary,array}
\usepackage[]{changes}

\usepackage{tikz}
\usetikzlibrary{arrows.meta,positioning,shapes.geometric}

\newcommand*{\rom}[1]{\expandafter\@slowromancap\romannumeral #1@}\makeatother

\usepackage[style=numeric-comp,sorting=none,
maxcitenames=3, maxbibnames=4]{biblatex}

\bibliography{bibl.bib}

\title{VISION-ICE: \underline{V}ideo-based \underline{I}nterpretation and \underline{S}patial \underline{I}dentification of Arrhythmia \underline{O}rigins via \underline{N}eural Networks in Intracardiac Echocardiography}

\author{\IEEEauthorblockN{Dorsa EPMoghaddam\IEEEauthorrefmark{1}, 
Feng Gao\IEEEauthorrefmark{3}, Drew Bernard\IEEEauthorrefmark{3} , Kavya Sinha\IEEEauthorrefmark{3}, 
Mehdi Razavi\IEEEauthorrefmark{4}, Behnaam Aazhang\IEEEauthorrefmark{2}}

\IEEEauthorblockA{\IEEEauthorrefmark{1}\IEEEauthorrefmark{2}Department of Electrical and Computer Engineering,
Rice University, United States of America}
\IEEEauthorblockA{\IEEEauthorrefmark{3}Electrophysiology Clinical Research and Innovations, Texas Heart Institute, United States of America }
\IEEEauthorblockA{\IEEEauthorrefmark{4}Department of Cardiology, Texas Heart Institute, United States of America}

\thanks{\IEEEauthorrefmark{1}Corresponding author: Dorsa EPMoghaddam (de11@rice.edu)} }

\begin{document}
\maketitle
 
\section*{Abstract}

Contemporary high-density mapping techniques and preoperative CT/MRI remain time and resource intensive in localizing arrhythmias. AI has been validated as a clinical decision aid in providing accurate, rapid real-time analysis of echocardiographic images. Building on this, we propose an AI-enabled framework that leverages intracardiac echocardiography (ICE), a routine part of electrophysiology procedures, to guide clinicians toward areas of arrhythmogenesis and potentially reduce procedural time. Arrhythmia source localization is formulated as a three-class classification task, distinguishing normal sinus rhythm, left-sided, and right-sided arrhythmias, based on ICE video data. We developed a 3D Convolutional Neural Network trained to discriminate among the three aforementioned classes. In ten-fold cross-validation, the model achieved a mean accuracy of $66.2\%$ when evaluated on four previously unseen patients (substantially outperforming the $33.3\%$ random baseline). These results demonstrate the feasibility and clinical promise of using ICE videos combined with deep learning for automated arrhythmia localization. Leveraging ICE imaging could enable faster, more targeted electrophysiological interventions and reduce the procedural burden of cardiac ablation. Future work will focus on expanding the dataset to improve model robustness and generalizability across diverse patient populations.

\textbf{Keywords:}
intracardiac echocardiography (ICE), 3D Convolutional Neural Network, arrhythmia localization, cardiac ablation, ICE video analysis

\section{Introduction}
Echocardiography is a widely used imaging modality that provides real-time visualization of cardiac structure and function across diverse clinical settings, from heart failure evaluation and valvular disease diagnosis to emergency care and perioperative monitoring \cite{mitchell2019guidelines, raissi2025contemporary}. This ultrasound-based, radiation-free technique operates by emitting high-frequency sound waves through a transducer probe, which receives reflected echoes to generate dynamic cardiac images.

Multiple echocardiographic modalities have been developed for different clinical contexts. Transthoracic echocardiography (TTE), the most common approach, acquires images externally through the chest wall. Transesophageal echocardiography (TEE) provides higher-resolution imaging by positioning the probe within the esophagus, close to posterior cardiac structures, and is especially valuable for valve and atrial evaluation \cite{hahn2013guidelines}. In contrast, intracardiac echocardiography (ICE) is an invasive technique performed in the catheterization laboratory or electrophysiology suite, where an ultrasound catheter inserted into the cardiac chambers delivers high-resolution images from within the heart itself \cite{hijazi2009intracardiac}. ICE has become indispensable for guiding interventions such as transcatheter valve repair or replacement, atrial septal defect closure, and atrial fibrillation ablation \cite{hijazi2009intracardiac, jongbloed2005clinical}. Together, these non-invasive and invasive approaches establish echocardiography as a cornerstone of diagnostic evaluation and image-guided therapy.


To address these limitations, there is growing interest in leveraging artificial intelligence (AI) and machine learning (ML) to augment echocardiography \cite{sakamoto2025artificial, sahashi2025ai}. Crucially, AI complements rather than replaces human expertise, enhancing efficiency and clinical decision-making \cite{sakamoto2025artificial}. Recent advances in machine learning, particularly deep learning, have enabled a broad spectrum of echocardiographic applications that were once only conceptual; ranging from view classification and image segmentation to automated quantification and disease prediction \cite{sakamoto2025artificial,sahashi2025ai,zhu2025automated}. Recent studies show that deep learning can rapidly estimate parameters such as ejection fraction and even anticipate cardiac disease or future outcomes directly from echocardiographic images, bypassing manual measurements \cite{sahashi2025ai}. Automated view classification models now reliably recognize standard echocardiographic views \cite{madani2018fast}, while segmentation networks such as U-Net delineate chambers and valves with high precision, supporting quantification of ventricular volumes and ejection fraction \cite{moradi2019mfp}. Critically, deep learning models can estimate left ventricular EF (LVEF) directly from raw echo cine loops with accuracy approaching expert human readings \cite{akan2025viviechoformer, ouyang2020video}. Beyond routine measurements, these models can identify pacemaker leads, detect ventricular hypertrophy, and even predict patient attributes such as age and sex, revealing latent phenotypic features beyond human perception \cite{ouyang2020video}. Machine learning has also advanced the evaluation of structural heart disease, improving detection of hypertrophic cardiomyopathy and valvular lesions by recognizing subtle echo patterns \cite{zhang2018fully}.

Among different deep learning models, Convolutional Neural Networks (CNNs) remain central to these applications, and are particularly well-suited to the spatiotemporal complexity of echocardiographic videos, excelling in tasks like view classification through hierarchical feature extraction. For example, Ouyang et al. \cite{ouyang2020video} developed a video-based deep learning CNN-based model that estimated LVEF on a frame-by-frame basis, effectively analyzing cardiac function at the level of individual heartbeats. Such models capture subtle variations in contractility that single-frame analyses might miss, with performance comparable to expert interpretation of ejection fraction and wall motion \cite{sakamoto2025artificial}.

Encoder–decoder architectures such as U-Net achieve high-precision chamber segmentation \cite{huang2022segmentation, penso2021automated}. Recurrent neural networks (RNNs), particularly Long short-term memory (LSTMs), can capture the temporal dynamics of echocardiography. Coupled with CNNs, they model motion across frames. Multimodal approaches are also gaining traction. Goto et al. \cite{goto2021artificial} introduced a model that jointly analyzes echocardiographic images and ECG signals, with separate networks for each modality whose outputs are fused for final prediction. Applied to amyloidosis detection, this cross-modal strategy demonstrated strong performance, highlighting the value of integrating heterogeneous data sources \cite{sakamoto2025artificial}.

Another recent development is the rise of vision–language foundation models that align echocardiographic images with expert textual interpretations \cite{sakamoto2025artificial, christensen2024vision}. For example, EchoCLIP, trained on over one million echo videos paired with reports, achieved state-of-the-art performance in estimating LVEF (MAE 7.1\%), identifying intracardiac devices (AUC up to 0.97), and enabling cross-modal retrieval. Its long-context variant, EchoCLIP-R, extended these capabilities to patient identification across studies and recognition of clinical transitions such as transplantation or surgery \cite{christensen2024vision}. Early studies also explore large language models (LLMs) for automated reporting or interactive query–response on echo findings \cite{alber2025medical, christensen2024vision}. While still in early stages, these vision–language models open the door to more interactive and explainable AI in echocardiography.

Improving model transparency remains critical for clinical adoption of AI in echocardiography. Explainable AI (XAI) techniques, such as Gradient-weighted Class Activation Mapping (Grad-CAM) and Shapley Additive Explanations (SHAP) provide visualizations of the regions most influential in a model’s decision \cite{selvaraju2017grad, huff2021interpretation}. By linking advanced machine learning with clinical insight, XAI enables safer and more interpretable AI-assisted echocardiography \cite{sakamoto2025artificial, sahashi2025ai}.

Although echocardiography has been extensively studied in recent years, research on intracardiac echocardiography remains limited, largely due to its invasive nature and lack of standardized, public datasets. Yet ICE provides high-resolution, real-time imaging crucial for guiding structural and electrophysiological procedures. Developing AI tailored to ICE could enhance procedural safety, efficiency, and diagnostic accuracy.

Limited studies have focused on ICE, though recent research is expanding this area. A very recent $2025$ study \cite{gungor2025fully} developed a fully automated deep learning algorithm to detect anatomic structures from intracardiac echocardiography images collected from the right atrium. The study used ICE images from $605$ electrophysiology procedures across $2$ institutions, achieving correct identification of $15$ out of $21$ anatomic structures with $>70\%$ precision and recall  \cite{gungor2025fully}. Another work evaluated a deep learning algorithm that creates $3$D left atrial anatomical rendering from ICE during atrial fibrillation ablation \cite{di2023feasibility}. The study included $28$ patients and demonstrated feasibility of generating $3$D shells of LA anatomy and structures compared to cardiac CT \cite{di2023feasibility}. Additionally, emerging research has explored multimodal approaches, such as a $2025$ study that developed deep learning methods for real-time pose estimation of ICE probes in $2$D X-ray fluoroscopy, representing progress toward ICE-XRF fusion systems that could enhance catheter navigation and positioning accuracy during interventional procedures \cite{severens2025toward}.

Despite these technological strides, a critical gap remains in leveraging ICE for arrhythmia localization, a fundamental challenge in electrophysiology that directly impacts ablation success rates and procedural efficiency. Current electrophysiology practice relies heavily on operator expertise and time-intensive mapping procedures to identify arrhythmogenic substrates. The development of AI-assisted arrhythmia localization systems could fundamentally transform this paradigm by providing objective, consistent guidance for identifying regions of interest during ablation procedures.

This study addresses this unmet clinical need by developing a novel deep learning framework specifically designed for ICE-based arrhythmia localization \cite{esmaeilpourmoghaddam2025po}. We formulate the problem as a three-class classification task, distinguishing between two anatomically distinct regions of interest and a control class, a clinically relevant yet computationally tractable approach that establishes the foundation for more sophisticated localization algorithms.

Our contribution is threefold: First, we introduce the first machine learning approach specifically tailored to arrhythmia localization using ICE imaging, addressing a previously unexplored application domain with significant clinical implications. Second, we present a carefully curated and annotated dataset designed for this specific task, providing a valuable resource for future research in ICE-based electrophysiology applications. Third, we demonstrate how specialized deep learning architectures can be adapted to the unique characteristics of ICE video data, potentially enhancing the precision and efficiency of ablation procedures.
The clinical significance of this work extends beyond technical innovation. Such systems have the potential to support ICE interpretation by providing expert-level decision assistance, which may be particularly valuable in centers with limited electrophysiology expertise, and could achieve broader clinical impact as access to AI deployment infrastructure continues to expand. This research represents an initial yet crucial step toward comprehensive AI-guided electrophysiology, where real-time ICE analysis directly informs therapeutic decision-making and optimizes patient outcomes.

\section{Methods}
\subsection{Dataset} 
Patients undergoing elective electrophysiology study procedures at Baylor St. Luke’s Hospital were assessed for eligibility. All patients provided signed informed consent for procedures. Only unidentifiable ultrasound images and associated electrocardiogram strips were collected. Patients were included if inclusion criteria were met: use of intracardiac echocardiography during procedure, use of a decapolar coronary sinus catheter, sinus rhythm at end of procedure with a ventricular rate of less than $100$ beats per minute, and ultrasound images of sufficient quality and without significant imaging artifact. 

Upon meeting inclusion criteria, baseline intracardiac ultrasound views were collected for $10$ beats using a Soundstar ultrasound catheter (Johnson and Johnson, New Brunswick, NJ, USA) and General Electric Vivid S$70$ machine (General Electric, Boston, MA, USA). Four standard views were collected: crista terminalis (CT), left pulmonary veins (LPV), mitral valve (MV), and tricuspid valve (TV). After baseline views were obtained, pacing was done at $600$ ms cycle length from DECANAV catheter (Johnson and Johnson, New Brunswick, NJ, USA) in the coronary sinus using electrodes $1-2$ (distal pacing). Upon stable capture, the four standard views were collected. Then, pacing was conducted at $600$ ms cycle length from in the coronary sinus using electrodes $9-10$ (proximal pacing). In instances where proximal $9-10$ had inadequate capture, electrodes $7-8$ were used. Four standard views were collected in similar fashion. 

A total of $39$ patients were included in the study. Table~\ref{tab:heartbeat_counts} provides a summary of the number of heartbeats collected from each patient.

\begin{table}[htbp]
\centering
\caption{Number of heartbeats collected for each patient in the dataset.}
\label{tab:heartbeat_counts}
\begin{tabular}{cc|cc}
\hline
\textbf{Patient ID} & \textbf{\# heartbeats} & \textbf{Patient ID} & \textbf{\# heartbeats} \\
\hline
P1 & 95 & P21 & 108 \\
P2 & 90 & P22 & 165 \\
P3 & 114 & P23 & 164 \\
P4 & 108 & P24 & 168 \\
P5 & 108 & P25 & 96  \\
P6 & 108 & P26 & 165 \\
P7 & 100 & P27 & 107 \\
P8 & 162 & P28 & 162 \\
P9 & 108 & P29 & 108 \\
P10 & 98  & P30 & 108 \\
P11 & 157 & P31 & 103 \\
P12 & 66  & P32 & 108 \\
P13 & 107 & P33 & 167 \\
P14 & 108 & P34 & 167 \\
P15 & 101 & P35 & 168 \\
P16 & 107 & P36 & 168 \\
P17 & 34  & P37 & 168 \\
P18 & 152 & P38 & 106 \\
P19 & 107 & P39 & 160 \\
P20 & 106 & & \\
\hline
\end{tabular}
\end{table}

\subsection{Preprocessing}

\begin{table*}[h!]
    \centering
    \includegraphics[width=0.9\textwidth]{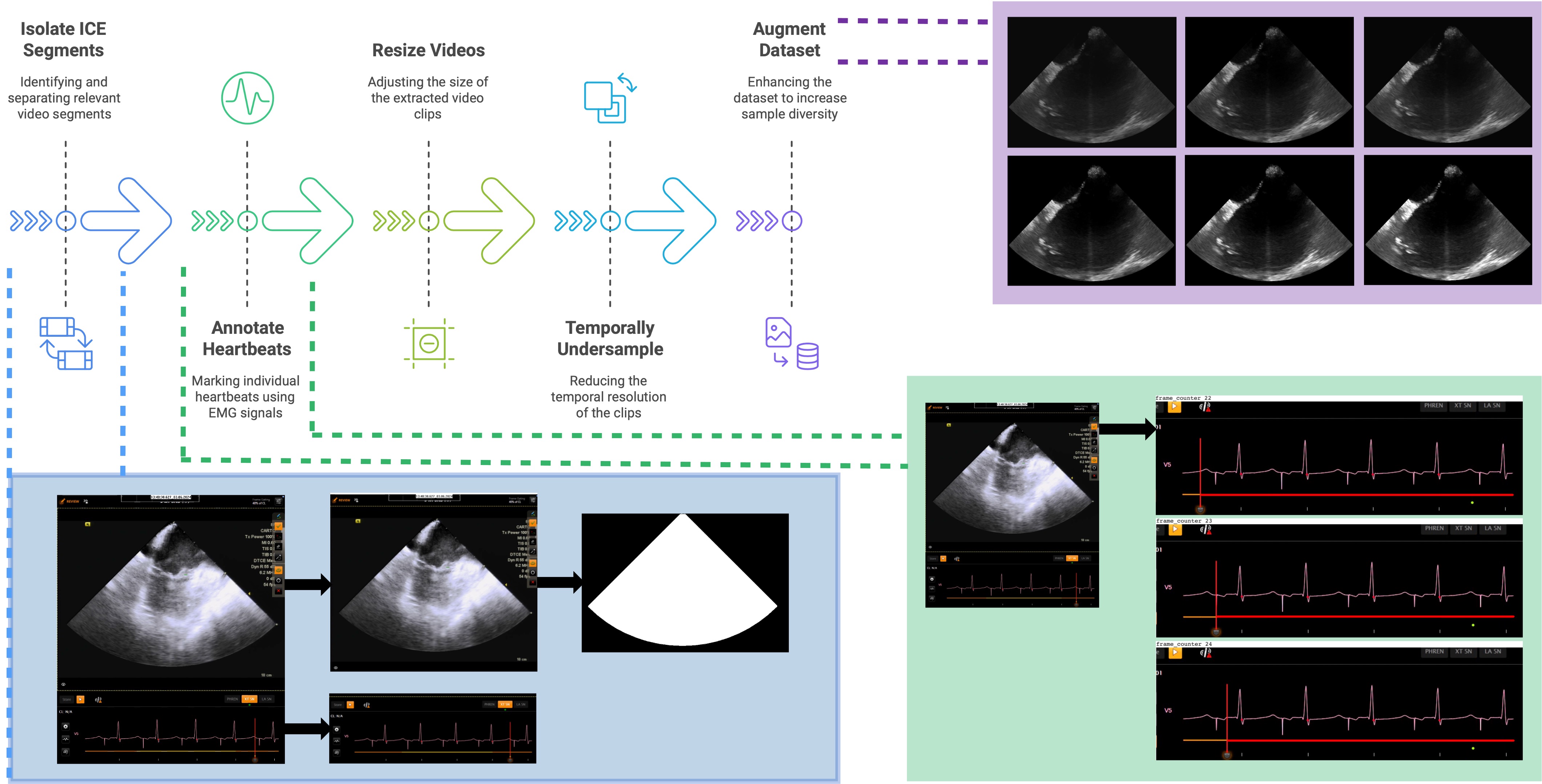}
     \captionof{figure}{Overview of the preprocessing pipeline. It includes masking of irrelevant regions, temporal segmentation, spatial cropping and resizing, and data augmentation.} \label{fig:preprocess}
\end{table*}

The preprocessing pipeline comprises several carefully designed stages to ensure data quality and integrity. It begins with identifying the ICE mask to isolate the region of interest and remove irrelevant background elements. Next, pixel intensities are normalized to the range $[0,1]$ to ensure a consistent input scale across all samples.  The third stage focuses on temporal segmentation, in which each recording is decomposed into individual cardiac cycles (i.e., heartbeats). This segmentation is guided by expert annotations provided by the clinical team, who examine the corresponding single-lead electrocardiogram signals to mark precise heartbeat boundaries. Both the PR segment and the end of each heartbeat were carefully annotated to establish accurate temporal boundaries. Each heartbeat is now treated as an individual sample in the next processing steps.

Subsequently, spatial cropping and resizing are applied to reduce data size and computational cost. The initial data has dimensions of $T \times 708 \times 1016$, where $T$ denotes the number of time frames, which varies across heartbeat samples. Cropping is applied to remove masked, non-informative regions that add computational burden without contributing diagnostic value, resulting in data of size $T \times 553 \times 756$. This step eliminated redundant areas while preserving clinically relevant content.   

At this stage, each sample was represented as a grayscale video tensor of shape $(1,T,H,W)$, with $H (553)$ and $W(756)$ denoting height and width. For temporal consistency, the number of frames was standardized to $32$. Sequences longer than $32$ frames were truncated, while shorter ones were padded by replicating the last frame until the desired length was reached. After preprocessing, all inputs had the standardized form $(1,32,H,W)$. To further reduce dimensionality, frames could be resized by a tunable factor through downsampling, maintaining coarse spatial structure while substantially lowering memory and computational demands during training and inference.

To enhance model robustness and generalizability, as well as to mitigate data scarcity, data augmentation was applied exclusively to the training set. For ICE recordings, the augmentation strategy included random brightness and contrast adjustments, random frame dropping, and the random addition of white Gaussian noise. For each original training sample, up to $A$ augmented variants were generated, where $A$ is a tunable parameter controlling the degree of dataset expansion. 

Specifically, for brightness and contrast jittering, given an input video frame $x \in [0,1]$, the brightness factor was sampled as $b \sim U[0.7,1.3]$ and the contrast factor as $c \sim U[0.5,1.5]$. The transformed input was then defined as: 

\[
x' = (x \cdot b - \mu)c + \mu, \quad \mu = \mathrm{mean}(x \cdot b),
\]
with the output clipped to the range $[0,1]$. The brightness and contrast factors were held constant across all frames of each video sample. 

Throughout the entire pipeline, special attention is given to preventing data leakage. All preprocessing steps are implemented in a manner that ensures no information from the test set contaminates the training process.

\begin{table*}[h!]
    \centering
    \includegraphics[width=0.7\textwidth]{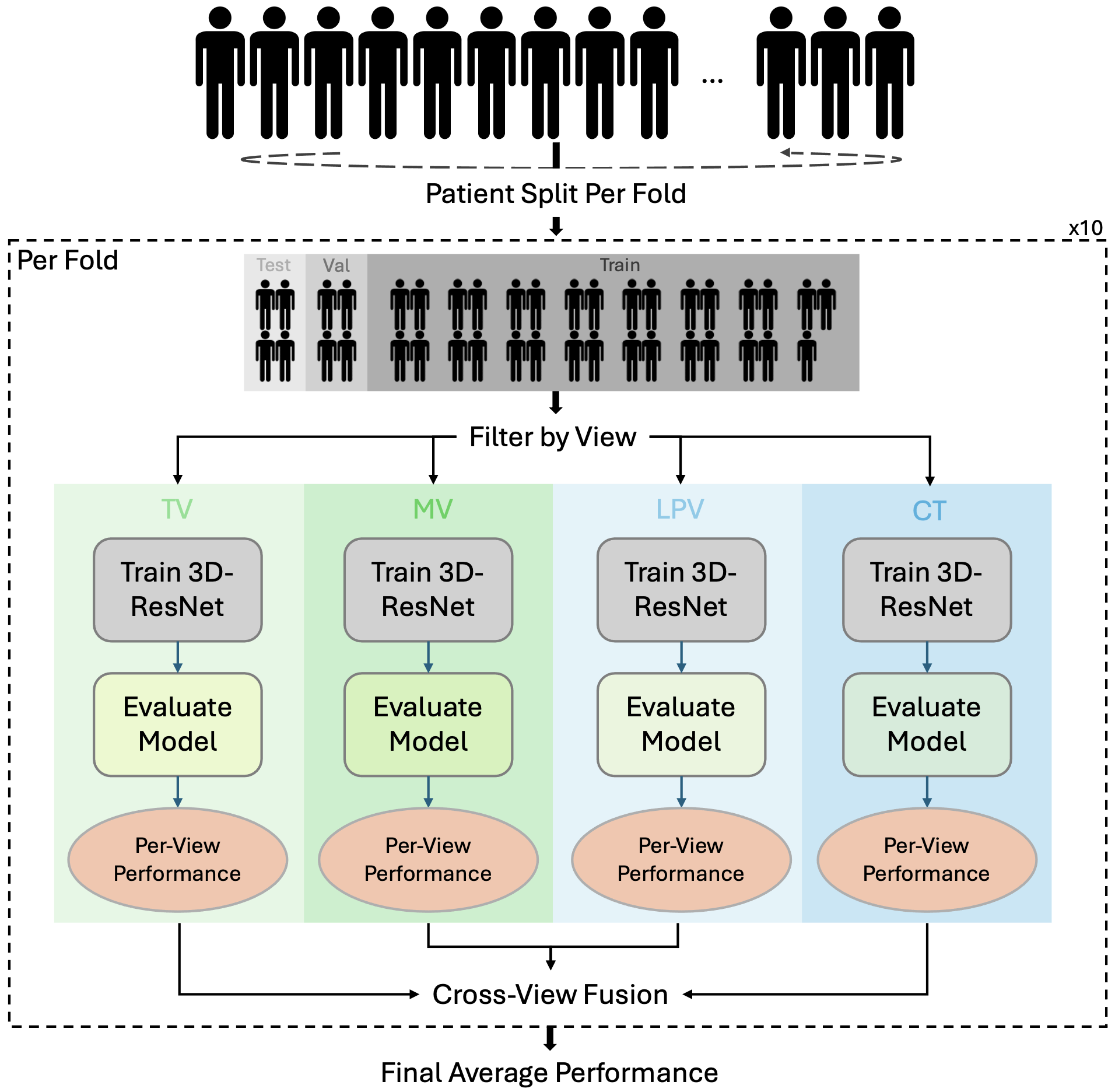}
     \captionof{figure}{Overview of the hierarchical evaluation framework for ICE-based arrhythmia localization. For each of the ten folds, patients are split into independent training, validation, and test subsets to prevent data leakage. Within each fold, four models are trained independently, one per anatomical view, using early stopping based on sample-level metrics. Clip-level predictions are then obtained by majority voting across heartbeats within each clip. At the final fusion stage, predictions from all available views are aggregated by majority voting to produce cross-view patient-level decisions. Final performance metrics are computed and averaged across folds to yield the overall evaluation summary.} \label{fig:flow}
\end{table*}

\subsection{Data Division}  

To ensure rigorous evaluation and prevent patient-level data leakage, we adopted a strictly inter-patient data splitting strategy. We use the term split to refer to a single partitioning of the patient cohort into three disjoint subsets: training, validation, and test. We use the term fold to refer to one complete iteration of the cross-validation procedure, which consists of: $(1)$ creating a unique split, $(2)$ training a model on the training subset, $(3)$ selecting the best model using the validation subset, and $(4)$ evaluating performance on the test subset. By employing multiple folds, we obtain robust performance estimates despite the limited cohort size, as each patient serves as validation or test data across different folds while maintaining strict separation within any individual fold.

Specifically, we employed $10$-fold patient-level cross-validation over $N = 39$ unique patients. Patients were indexed using a fixed, deterministic ordering based on anonymized identifiers. In each fold $r$, we assigned patients to subsets using a circular sliding window: four consecutive patients were allocated to the test set, the subsequent four to the validation set, and the remaining $31$ patients to the training set. This yielded an approximate $80/10/10$ patient-level split per fold and ensured that every patient appeared in both validation and test sets across the $10$ folds, while never appearing in more than one subset within the same fold. The training set was used for model fitting, the validation set guided hyperparameter tuning and checkpoint selection, and the test set provided final performance estimates to assess generalization to unseen patients. Figure~\ref{fig:flow} illustrates this data division scheme.

\subsection{Model Architecture}  

The task was formulated as a three-class video classification problem, comprising a control class and two classes corresponding to distinct arrhythmia origins. As these arrhythmias were surgically induced through controlled pacing, the terms 'arrhythmia origin' and 'pacing site' are used interchangeably throughout this manuscript. 

Each sample is annotated with a pacing label and a view label. Pacing labels comprise: normal sinus rhythm (NSR, class id $0$), distal coronary sinus pacing (DIST, id $1$), and proximal coronary sinus pacing (PROX, id $2$). View labels comprise: tricuspid valve (TV, id $0$), mitral valve (MV, id $1$), left pulmonary vein (LPV, id $2$), and crista terminalis (CT, id $3$). 

We adopted a  pre-trained $3$D ResNet-$18$ backbone (\texttt{torchvision r3d\_18}) with a custom single-channel input stem. A 3D convolutional adapter maps the grayscale input to 64 channels using a $(9\times7\times7)$ kernel, stride $(1,3,3)$, and padding $(1,3,3)$, followed by batch normalization and 3D dropout (dropout probability $p = 0.1$). The original stem convolution was removed and replaced by this adapter. After the ResNet backbone, we applied dropout ($p=0.2$) and a fully connected layer producing logits for the three classes.

\begin{table*}[h!]
\centering
\small
\begin{tabular}{c|c|c|c|c|c}
\hline
\textbf{Fold} & \textbf{TV} &  \textbf{MV} &  \textbf{LPV} &  \textbf{CT} &  \textbf{Cross-View Majority Voting} \\
\hline
1  & 72.73  & 90.90 &  63.63  & 54.54 & 72.73 \\ 
2  & 75.00  & 83.33 &  66.67  & 50.00 & 75.00 \\ 
3  & 75.00  & 91.67 &  83.33  & 81.81 & 91.67 \\ 
4  & 66.67  & 66.67 &  66.67  & 58.33 & 75.00 \\ 
5  & 70.00  & 80.00 &  100.00 & 80.00 & 90.00 \\ 
6  & 70.00  & 58.33 &  50.00  & 41.66 & 66.67 \\ 
7  & 75.00  & 91.67 &  91.67  & 75.00 & 83.33 \\ 
8  & 66.67  & 75.00 &  66.67  & 58.33 & 66.67 \\
9  & 75.00  & 66.67 &  58.33  & 60.00 & 75.00 \\ 
10 & 66.67  & 66.67 &  83.33  & 70.00 & 66.67 \\ 
\hline
\textbf{Mean} & \textbf{71.27} & \textbf{77.09}& \textbf{73.03} & \textbf{62.97}& \textbf{76.27} \\
\hline
\end{tabular}
\caption{Validation accuracies across $10$ folds for individual ICE views (computed using clip-level majority voting) and for the combined cross-view majority voting outcome. }
\label{tab:cv_crossview_rounds_val}
\end{table*}

\begin{table*}[h!]
\centering
\small
\begin{tabular}{c|c|c|c|c|c}
\hline
\textbf{Fold} & \textbf{TV} &  \textbf{MV} &  \textbf{LPV} &  \textbf{CT} &  \textbf{Cross-View Majority Voting} \\
\hline
1  & 50.00  & 75.00  & 75.00 & 50.00 &  83.33 \\ 
2  & 63.63  & 45.45  & 63.63 & 54.54 & 63.64  \\ 
3  & 58.33  & 66.67  & 41.67 & 50.00 & 66.67  \\ 
4  & 66.67  & 75.00  & 83.33 & 72.72 & 66.67  \\ 
5  & 66.67  & 58.33  & 66.67 & 58.33 & 66.67  \\ 
6  & 70.00  & 80.00  & 70.00 & 60.00 & 90.00  \\ 
7  & 40.00  & 50.00  & 50.00 & 50.00 & 41.67  \\ 
8  & 75.00  & 75.00  & 58.33 & 58.33 & 66.67  \\
9  & 66.67  & 66.67  & 66.67 & 33.33 & 58.33  \\ 
10 & 58.33  & 58.33  & 58.33 & 60.00 & 58.33  \\ 
\hline
\textbf{Mean} & \textbf{61.53} & \textbf{65.05}& \textbf{63.36} & \textbf{54.73}& \textbf{66.20} \\
\hline
\end{tabular}
\caption{Test accuracies across $10$ folds for individual ICE views (computed using clip-level majority voting) and for the combined cross-view majority voting outcome. }
\label{tab:cv_crossview_rounds}
\end{table*}

\subsection{Training Protocol}  

Models were trained with AdamW (learning rate $10^{-5}$, weight decay $10^{-3}$) and a class-weighted cross-entropy loss (uniform by default). Mixed precision (AMP) was enabled on GPU, and gradients were clipped to a global norm of $1.0$. Training proceeded for up to $150$ epochs (where one epoch denotes a full pass through the entire training dataset) with early stopping based on validation accuracy (patience $20$; batch size $8$). In one variant, a \texttt{ReduceLROnPlateau} scheduler halved the learning rate when validation loss plateaued.  

As an additional safeguard against overfitting, training was stopped if training accuracy exceeded $90\%$ while validation accuracy failed to improve for $10$ epochs. The checkpoint corresponding to the highest validation accuracy was saved and reloaded for final evaluation.  

For reproducibility, we fixed the random seeds in Python, NumPy, and PyTorch, and enabled deterministic implementations for operations such as convolution and pooling.

\subsection{Evaluation}  

Model performance was evaluated at three hierarchical levels to capture both fine-grained and aggregated behavior. Sample-level refers to predictions on individual heartbeat samples, clip-level aggregates predictions across all heartbeats within a single video clip, and patient-level combines results across all clips from the same subject, encompassing all four imaging views.


Performance was quantified using accuracy defined as
\begin{equation}
\text{Accuracy} = \frac{\text{number of correct predictions}}{\text{total number of predictions}}.
\end{equation}

For each of the ten folds, separate models were trained for the four anatomical views (TV, MV, LPV, CT). Sample-level metrics guided model selection and early stopping, while clip-level performance was obtained via majority voting across all heartbeats within a clip. Patient-level performance was then derived through cross-view fusion, where predictions from all available views were aggregated by majority voting. For each patient–class pair, the final label was the mode of the corresponding clip-level predictions, reducing intra-patient variability and mitigating sample-level noise. The evaluation workflow is illustrated in Figure~\ref{fig:flow}.

\begin{table*}[h!]
    \centering
    \includegraphics[width=0.9\textwidth]{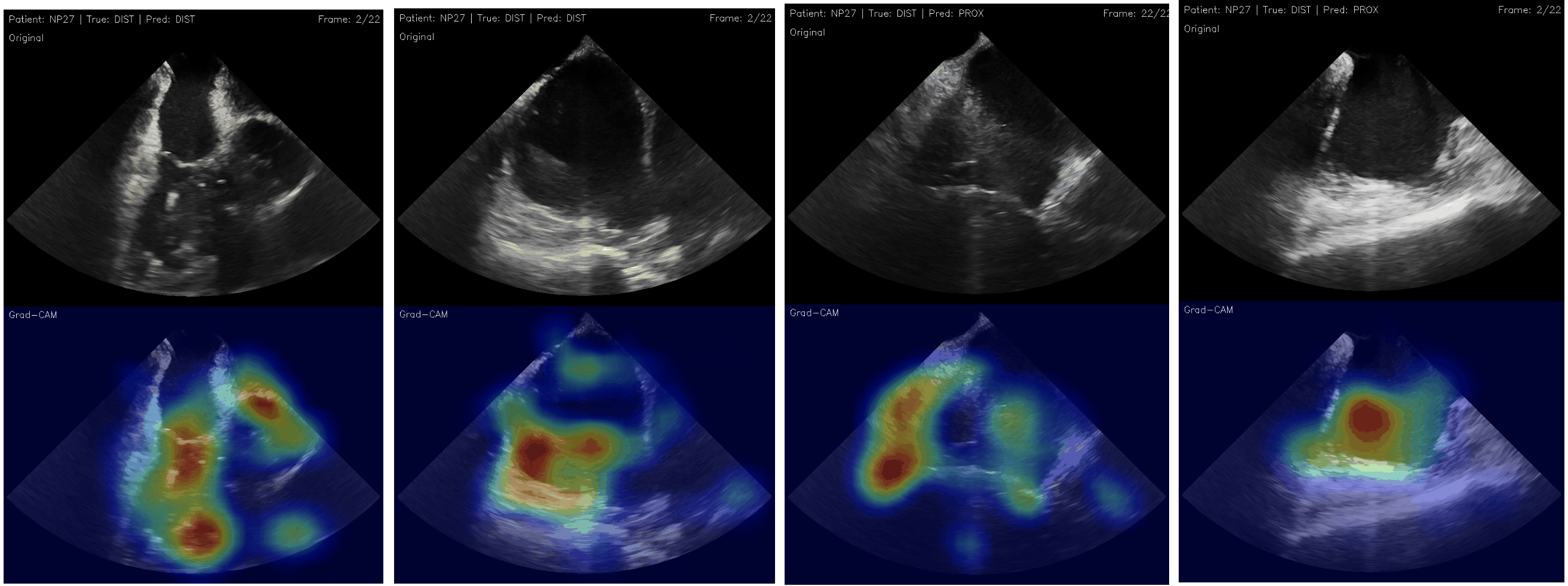}
     \captionof{figure}{Grad-CAM visualizations derived from the deepest convolutional layer (layer $4$) of the modified ResNet18-3D. The top row displays the original intracardiac echocardiography frames, and the bottom row shows the corresponding activation heatmaps. The two columns on the left correspond to correctly classified samples, and the two columns on the right illustrate misclassified cases. Warmer colors (red/yellow) indicate regions with higher contribution to the model’s prediction, while cooler colors (blue) represent areas of lower importance.} \label{fig:grad-cam}
\end{table*}

\subsection{Visualization: Grad-CAM and Attention Maps}

We implement $3$D Grad-CAM by registering hooks on the deepest available $3$D convolution within the network. Spatial-temporal gradients are globally averaged to obtain channel weights, which are linearly combined with activations, ReLUed, normalized, and upsampled trilinearly to the input resolution. For qualitative inspection, we overlay heatmaps on each frame and export side-by-side GIFs (“Original” vs “Grad-CAM”) annotated with patient ID, true label, predicted label, and frame index.

\section{Results}

We trained view-specific models on heartbeat samples from each anatomical view (TV, MV, LPV, CT). Clip-level classification performance via majority voting is reported in Tables~\ref{tab:cv_crossview_rounds_val} (validation) and~\ref{tab:cv_crossview_rounds} (test). Patient-level performance, obtained through cross-view majority voting to exploit complementary information and improve robustness, is shown in the final column of Tables~\ref{tab:cv_crossview_rounds_val} and~\ref{tab:cv_crossview_rounds}.

To further enhance interpretability, we applied Grad-CAM visualizations to highlight the spatial and temporal regions that contributed most to the model’s predictions. This approach enables qualitative verification that the model attends to clinically meaningful structures within the ICE recordings. Figure \ref{fig:grad-cam} presents an example in which Grad-CAM is applied to the last convolutional layer of our model, which has the largest receptive field and therefore provides high-level, class-specific evidence for the prediction.

\section{Discussion}
The results demonstrate that 3D-CNNs effectively capture the spatiotemporal dynamics of ICE recordings for arrhythmia classification. Fine-tuning pre-trained models achieved strong performance, underscoring the value of transfer learning in medical video analysis where annotated datasets are often limited. Data augmentation further improved robustness, although a moderate level of augmentation proved most effective.

Because of the way the dataset was constructed, class balance was maintained. The primary challenge, however, lies in learning effectively from a very limited number of patients, given the substantial inter-patient variability inherent in medical data. This limitation is further reflected in the variability observed across cross-validation folds. Due to the small cohort size and heterogeneous patient characteristics, different random splits can lead to markedly different distributions between training, validation, and test sets. Folds exhibiting higher performance likely correspond to splits in which these subsets are more similar in distribution, whereas lower-performing folds arise when the validation and test sets contain patients that are less well represented in the training data. This behavior highlights the sensitivity of model performance to data partitioning in low-sample, high-variability clinical settings. The observed gap between validation and test performance is expected and reflects the combined effects of inter-patient heterogeneity and limited sample size, with test accuracy providing a more conservative and unbiased estimate of generalization to truly unseen patients. 

Model interpretability is another critical consideration for clinical adoption. To address this, we incorporate Grad-CAM visualizations to highlight the spatial and temporal regions that influence model predictions, thereby enhancing clinical trust and transparency. As Figure~\ref{fig:grad-cam} demonstrates, the model’s attention aligns with physiologically relevant atrial structures, including the interatrial septum, mitral annulus region, and crista terminalis, suggesting meaningful feature learning rather than reliance on irrelevant cues. Differences in attention patterns across imaging views further indicate that certain views may provide more informative signals for atrial activation discrimination, motivating future work on view-specific optimization. Future work will also explore alternative architectures, such as Swin3D Transformers, as well as ensemble strategies to further improve accuracy and robustness.

\section{Conclusion}
We presented a deep learning framework for arrhythmia classification using ICE videos. Leveraging a model with careful preprocessing, augmentation, and subject-independent evaluation, the proposed method achieved promising results on a challenging task. This study demonstrates the feasibility of applying spatiotemporal deep learning to ICE and sets the stage for future advancements in AI-assisted electrophysiology. Ongoing work focuses on expanding the dataset, integrating interpretability methods, and exploring domain adaptation across different echocardiographic modalities.

\section*{Acknowledgments}
This research was funded by the National Heart Lung and Blood Institute (grant R01HL144683). Data were collected under Institutional Review Board approval from Baylor College of Medicine and Affiliated Hospitals (protocol H-43925, approved September 12, 2018).

\section*{Conflict of interest statement}
The authors of this paper have no conflicts of interest to report.

\printbibliography

\end{document}